\documentclass{article}

\usepackage{neurips_2023}
\usepackage{amsmath}
\usepackage{graphicx}
\usepackage{pgfplots}
\usepackage{booktabs} % For better-looking tables
\usepackage{caption}  % For captions
\pgfplotsset{width=8cm, compat=1.9}

\usepackage[utf8]{inputenc} % allow utf-8 input
\usepackage[T1]{fontenc}    % use 8-bit T1 fonts
\usepackage{hyperref}       % hyperlinks
\usepackage{url}            % simple URL typesetting
\usepackage{booktabs}       % professional-quality tables
\usepackage{amsfonts}       % blackboard math symbols
\usepackage{nicefrac}       % compact symbols for 1/2, etc.
\usepackage{microtype}      % microtypography
\usepackage{xcolor}         % colors
\usepackage{float}

\title{Stronger Graph Transformer with Regularized Attention Scores}

\author{
  Eugene Ku\\
  HDSI, UCSD\\
  euku@ucsd.edu\\
}

\begin{document}

\maketitle

\begin{abstract}
    \hspace{1cm}Graph Neural Networks are notorious for its memory consumption. A recent Transformer-based GNN called Graph Transformer is shown to obtain superior performances when long range dependencies exist. However, combining graph data and Transformer architecture led to a combinationally worse memory issue. We propose a novel version of "edge regularization technique" that alleviates the need for Positional Encoding and ultimately alleviate GT's out of memory issue. We observe that it is not clear whether having an edge regularization on top of positional encoding is helpful. However, it seems evident that applying our edge regularization technique indeed stably improves GT's performance compared to GT without Positional Encoding. 
\end{abstract}

% \begin{center}
% \end{center}

\section{Introduction}
\hspace{1cm}Graph Neural Networks (GNNs) have emerged as a powerful tool for learning representations of data structured as graphs, finding applications in diverse fields from social network analysis to molecular biology. Among all GNNs, the most popular variant is Message Passing Neural Networks (MPNN), namely Graph Convolution Network [1]. This MPNN architecture utilizes the graph structure by sending messages to only its neighbors, which allows MPNN to enjoy a linear time complexity with respect to the number of edges in the graph. 

\hspace{1cm}However, these MPNN architectures have theoretical weaknesses that are so-called oversquashing [2] and oversmoothing [3]. Oversquashing occurs when information that are k-hops away get exponentially weaker as the reception field of a node grows exponentially ($degree^k$). One can also take a perspective that as information/message flows from one node to another, it gets "corrupted" by other messages at the rate of the degree of the node. Oversmoothing happens when non-expressive aggregation pooling occurs frequently (namely sum, mean, max, etc). These aggregations are known to make nodes more similar to its neighbors. After a certain number of aggregations, nodes features will be indistinguishable from that of other nodes, which significantly degrades our expressiveness and performance. Overall, these oversquashing and oversmoothing problems hurt MPNNs ability to build a deeper GNN that capture long range dependencies present in the graph. 

% insert oversquashing image here

\begin{figure} [htbp]
   \centering
   \includegraphics[width=0.5\textwidth,height=5cm]{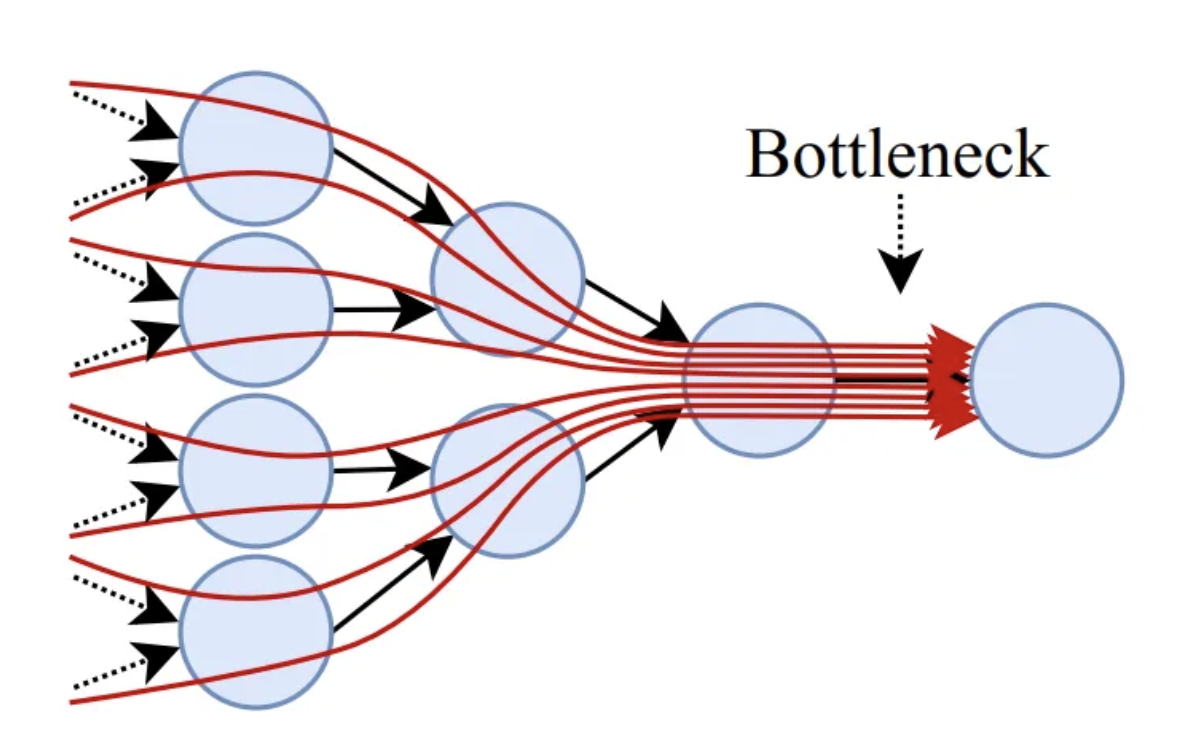}
   \caption{Oversquashing [2]}
\end{figure}

% \footnotetext{* Corresponding author}
% \footnotetext{github: \url{https://github.com/Eugene29/GraphGPS_Edge_Regularization}}
\begingroup
\renewcommand\thefootnote{}\footnotetext{Github code: \url{https://github.com/Eugene29/GraphGPS_Edge_Regularization}}
\endgroup

% However, as the complexity and size of graph-structured data grow, traditional GNNs, particularly Message Passing Neural Networks (MPNNs), face limitations in scalability and efficiency.

% This project introduces the GraphGPS model, a novel architecture that addresses these limitations. The GraphGPS model stands out for its ability to efficiently process large-scale graphs, capturing both local and global structural information. 

\hspace{1cm}However, new novel approach has been suggested to overcome these limitations. Transformers [4], who are known for capturing long range dependencies well as seen in Natural Language Processing (NLP), also seem to capture the long range dependencies well in Graphs. An architecture called GraphGPS [10] uses a hybrid approach of utilizing MPNNs and GTs that intuitively makes up for each architectures weakness such as weak inductive bias (GT) and poor long range capability (MPNN). 

\hspace{1cm}Despite the benefits, GraphGPS especially suffers from having excessive memory usage during training, and therefore, cannot be trained on graphs bigger than couple of thousands nodes. To alleviate this problem of GraphGPS and GT, we propose a method to optimize the computation of these architectures by feeding Graph's structure to our architecture without having to add any extra feature/positional encoding. Recently, similar approach to ours has taken place by the name of "edge regularization" [5]. However, ours differs in how we incorporate backpropogation cutoff technique and how we compute the loss function (section 3). We hypothesize that GT with our regularization technique would train in a more stable manner without having to add positional encodings that make our data orders of magnitude larger.

% By integrating the strengths of MPNNs and Graph Transformers (GTs), GraphGPS provides a unique solution to the challenge of long-range dependency modeling in graphs.

% The primary objective of this project is to improve upon the efficacy and scalability of the GraphGPS model by adding inductive bias to GT architecture in a way that does not increase the number of features. We aim to show that GraphGPS not only outperforms existing methods in standard benchmarks. 

% This is particularly important for tasks where understanding the global structure of the graph is as crucial as understanding local neighborhoods.

\section{Background \& Related Work}

\subsection{Evolution of Graph Neural Networks}

\hspace{1cm}The evolution of Graph Neural Networks (GNNs) marks a significant trajectory in the field of deep learning. Initially, neural networks were designed for grid-like data structures such as images and text. However, the complexity of real-world data, often represented as graphs, demanded more specialized models. This led to the development of the first GNN models, which aimed to capture the relational information inherent in graph structures.

% The early GNNs, while innovative, faced challenges in terms of scalability and efficiency. They were primarily designed for small graphs and struggled with larger, more complex structures. This limitation spurred further research, leading to the development of more advanced models like Graph Convolutional Networks (GCNs) and later, Message Passing Neural Networks (MPNNs). 
\hspace{1cm}These models enhanced the ability to learn from graph data but still had constraints, particularly in handling long-range dependencies within graphs. This backdrop set the stage for the emergence of Graph Transformers, a new class of models that addressed some of these longstanding issues. By incorporating attention mechanisms, the distances between all pairs of nodes suddenly became one, transforming the graph structure into a complete graph. Now, the problem of oversmoothing and oversquashing no longer existed. 

\subsection{Introduction to Graph Transformers}

\hspace{1cm}
% Graph Transformers have emerged as a solution to the limitations of traditional Message Passing Neural Networks, notably in handling long-range dependencies within graphs. 
Unlike MPNNs, which aggregate information locally, Graph Transformers learn relationships between nodes globally, irrespective of their position in the graph. This feature enables them to capture global graph structures more flexibly and effectively.

% \hspace{1cm}Central to Graph Transformers is the concept of self-attention, which allows each node to weigh the importance of other nodes' information dynamically. This process leads to a more flexible and comprehensive understanding of the graph's topology.

% *** show the table from Presentation that shows that models with GT performed much better. 

\begin{figure} [htbp]
   \centering
   \includegraphics[width=0.95\textwidth,height=4cm]{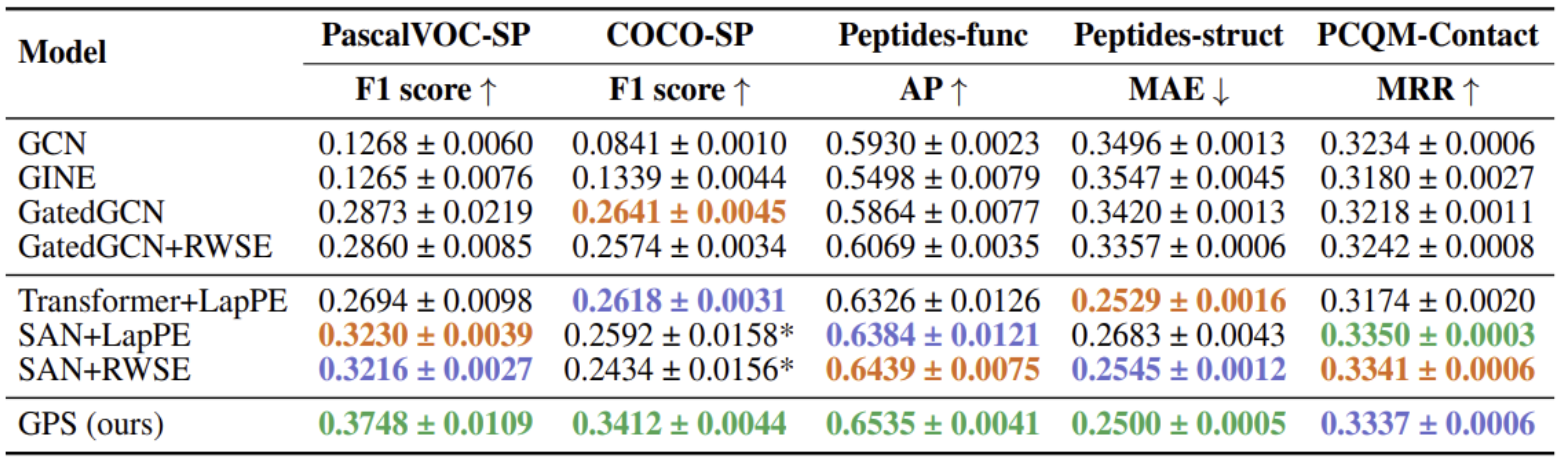}
   \caption{Superior Performance of models with Graph Transformers on LRGB[6]}
\end{figure}

\subsection{Limitations of Graph Transformers}

\hspace{1cm}However, utilizing Graph Transformer creates a new significant problem: our graph structure has disappeared. To make up for this disappearance of graph structure and inductive bias, many researchers have came up with ways to make up for these inductive bias through positional encoding that were successful in the NLP and Computer Vision (CV) field [7]. Most popular positional encodings include Laplacian Eigenmaps [8] and Random Walk Structural Encoding [9]. Notably, these popular positional encodings have problems like Laplacian Eigenmaps' sign ambiguity which creates $2^k$ combinations (although SignNet[11] seems to alleviate this problem). For Random Walk Structural Encoding, the structural encoding lacks the uniqueness that Laplacian Eigenmaps have. The visualizations of positional encodings are show in the figure 3, 4. A canonical positional encoding has yet been introduced. 

\hspace{1cm}On the bright side, these positional encoding allows both MPNN and GT to go beyond the past theoretical limitation of Color Refinement algorithm, which have been shown that if the Color Refinement Algorithm cannot distinguish two graphs, then MPNN certainly cannot. With the positional encoding, each nodes are now able to gain more information about their local structure and global position that they could go beyond the expressivity of vanilla message passing architecture. 

\hspace{1cm}However, on the dark side, addiing positional encoding poses \textbf{yet another} problem because Transformers in general are memory expensive due to its quadratic complexity. Positional encodings aggravate the memory problem.

% % Figures from “Graph Neural Networks with Learnable Structural and Positional Representations” by Dwivedi et. al.
% \begin{figure} [htbp]
\begin{figure}[H]
    \centering
    \includegraphics[width=0.9\linewidth]{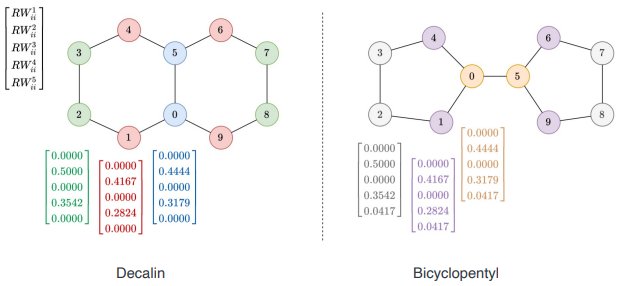}
    \caption{RWSE allows expressiveness beyond Color Refinement Algorithm [2]}
    \label{fig:enter-label}
\end{figure}
% \begin{figure} [htbp]
\begin{figure}[H]
    \centering
    \includegraphics[width=0.7\linewidth, height=5cm]{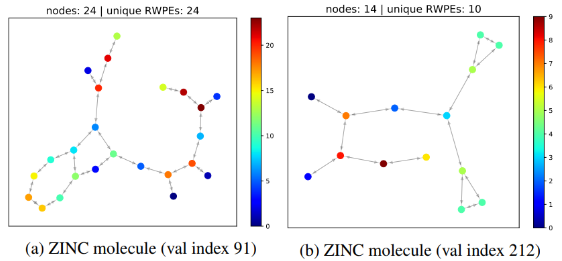}
    \caption{RWSE does not guaranteed to show unique encoding for each node[2]}
    \label{fig:enter-label}
\end{figure}

% This report will detail the design, implementation, and evaluation of the GraphGPS model. We begin by reviewing relevant literature and existing methods in graph neural networks to contextualize our work. Following this, we describe the proposed GraphGPS methodology, emphasizing its hybrid architecture and scalability features. We then present a comprehensive analysis of the results obtained from applying GraphGPS to various datasets. Finally, we discuss the limitations of our approach and suggest directions for future research, concluding with the implications of our findings in the broader context of graph representation learning.

% \subsection{Limitations of Message Passing Neural Networks}

% Message Passing Neural Networks (MPNNs) have been pivotal in the advancement of graph neural networks, allowing for effective learning of node representations by aggregating information from their neighbors. However, MPNNs have inherent limitations, especially in capturing global graph properties and long-range dependencies. Their reliance on local neighborhood information restricts their ability to learn broader graph contexts, which is crucial in many real-world applications.

% Furthermore, MPNNs face scalability challenges. As graph sizes increase, the computational and memory requirements of MPNNs grow significantly, making them less efficient for large-scale graphs. This scalability issue is particularly acute in tasks involving very large graphs, where the number of nodes and edges can be in the millions.

% \section{General formatting instructions}
\subsection{GraphGPS Architecture}\
\hspace{1cm} To evaluate the performance of our regularization technique, we chose a recently developed architecture called GraphGPS[10] which is a hybrid approach of GT and MPNN. The backbone architecture of our proposed method, GraphGPS, is as follows: 
\begin{equation*}
\begin{aligned}
\text{for } \ell &= 0, 1, \ldots, L-1 \\
&(a) \quad \tilde{X}^{\ell+1}_M, E^{\ell+1} \leftarrow \text{MPNN}^{\ell}_e (X^{\ell}, E^{\ell}, A) \\
&(b) \quad \tilde{X}^{\ell+1}_T \leftarrow \text{GlobalAttn}^{\ell} (\tilde{X}^{\ell}) \\
&(c) \quad X^{\ell+1}_M \leftarrow \text{BatchNorm} (\text{Dropout} (\tilde{X}^{\ell+1}_M) + X^{\ell}) \\
&(d) \quad X^{\ell+1}_T \leftarrow \text{BatchNorm} (\text{Dropout} (\tilde{X}^{\ell+1}_T) + X^{\ell}) \\
&(e) \quad X^{\ell+1} \leftarrow \text{MLP}^{\ell} (X^{\ell+1}_M + X^{\ell+1}_T)
\end{aligned}
\end{equation*}
\label{gen_inst}
\hspace{1cm}Intuitively, at every layer, the model diverges into two lanes and merges back at the end of each layer for a MLP layer. As a side note, from the authors' experience of using GraphGPS, the residual connections used at step (a), (b), (e) played a crucial role in performance, even more than the existence of positional encoding. 

\section{Proposed Method}
\hspace{1cm}Our proposed edge regularization technique is simple. At each GT layer, we cache the attention score matrix that have been computed from query and key matrix. Then, at the gradient calculation step, we calculate a additional loss function that takes the cached attention score matrix and the ground truth adjacency matrix as inputs. Further, we ensured that the gradients from the new loss function only affects the parameters of attention mechanism (i.e. computation of Q, K). 
% The authors implementation of backpropgation cut-off included storing the first gradients using the first forward step using the main loss function and then only storing the gradients of query matrix and key matrix at the second forward step. The optimizer.step() was only taken once per mini-batch. 

\hspace{1cm}The motivation of cutting off the backpropagation was to not disrupt the learning of useful node representation for minimizing the main loss function. 
% Empirically, we saw that not applying this backpopagation cut-off technique slightly degraded the model's performance. 
Lastly, we saw that applying the regularization on the softmaxed attention weight was fruitless. This was due to attention weights not being able to minimize the regularization loss well because of the constraints of Softmax (rows always sum up to 1). Instead, we applied a separate sigmoid function to our attention score matrix to make up for the constraint of the Softmax. 

% In mathematical terms:

% \begin{center}
%     $L_{\text{total}} = L + L_{\text{reg}}$

%     $L_{\text{reg}} = L_{chosen\_loss}(\text{Sigmoid}(A_{\text{score}}), \text{Adj Indices})$ \\
% \end{center}

% Here, $A_{score}$ is the attention score for each layer. $L_{reg}$ computes loss where adjacency indices are present. Loss Functions Explored: Cross-Entropy (CE), L1, L2, etc.\\

% Here is what we add to existing GraphGPS:
\hspace{1cm}The edge regularization loss we propose is as follows. Notations: Here, Q represents the query matrix while K represents the key matrix, where $d_{emb}$ is the length of the node feature.

\begin{equation*}
\begin{aligned}
\text{for } \ell &= 0, 1, \ldots, L-1 \\
    &1.\quad e_{ij} = QK^T / \sqrt{d_{emb}} \\
    &2.\quad L_{edge\_reg} {\scriptstyle+=} L1\_loss(Sigmoid(E), A) \quad \text{where E is the}
    \text{matrix of $e_{ij}$ and $A$ is adjacency matrix.}\\
\end{aligned}
\end{equation*}
\text{end for} \\
$\quad L_{total} = L_{main} + L_{edge\_reg}$
\label{gen_inst}

\hspace{1cm}The three dataset that we have chosen to test our proposed method are all from Long Range Graph Benchmark [6], which Graph Transformers are designed to perform well on. Out of all the collection, we chose to use Peptides-func, Peptides-Struct, and PascalVOC-SP. More details of our GraphGPS configuration is listed in the appendix.

\begin{figure} [H]
    \centering
    \includegraphics[width=0.6\linewidth]{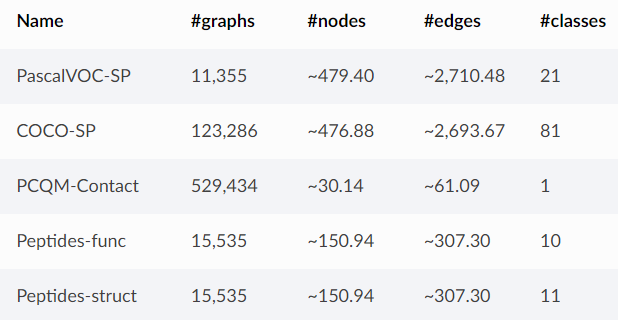}
    \caption{LRGB Dataset description [6]}
    \label{fig:enter-label}
\end{figure}

\section{Results}
\hspace{1cm} For results, we found that using Cross Entropy as our regularization hurts our models performance by a lot. Intuitively, the cross entropy would be too strict with sticking with the ground truth adjacency matrix as the loss would never allow full re-wiring ($a_{ij}=1$) between disconnected nodes. Interestingly, when observing the behaviour between having the backpropagation cut-off vs. allowing full backpropagation, we saw that allowing full backpropagation did better when the chosen metric was Cross Entropy. We unfortunately did not have the time to further investigate how the full backpropagation would do when L1 was chosen (future study). 

\hspace{1cm}We would like to acknowledge that we are not able to provide error bars for all measurements due to computation and time limitations (e.g. each run of PCQM-Contact's took 13+ hrs). We focused on providing the error bars on the baseline scores and on the best performing configurations (in first attempt.) Also note that the bar's values are the \textbf{first metric that we have obtained, and the error bar includes the values of two extra runs.} No error bar means, the metric was only ran once. Lastly, the y-values are zoomed-in in order to better depict the differences (most baselines are only 0.1 away from current State-of-the-art). 

% We see that without having the backpropgation cut off, the model improves by a non-siginficant margin. However, with the backpopagation cut off, our model saw a significant increase in GraphGPS' best performing score.
% [htbp]

\begin{figure}[H]
\begin{tikzpicture} 
\begin{axis}[
    ybar,
    title={Performance Effect of Regularization on Peptides-Struct \textbf{(w/ Positional Encoding)}},
    width = 13cm,
    enlarge x limits={true},
    yticklabel style={
    /pgf/number format/.cd,
    fixed,               % Use fixed point arithmetic (no scientific notation)
    fixed zerofill,      % Fill numbers with zeros to ensure uniform decimal places
    precision=4,         % Number of digits after the decimal point
    },
    ylabel={Mean Absolute Error},
    xlabel={Regularization Hyperparameters},
    symbolic x coords={
        No Reg, 
        0.01\, CE,
        0.05\, CE,
        0.1\, CE, 
        0.5\, CE, 
        w/out backprop cut-off 0.01\, CE,
        w/out backprop cut-off 0.05\, CE,
        0.01\, L1,
        0.05\, L1,
        0.1\, L1,
        0.5\, L1,
        },
    xtick=data,
    nodes near coords={\pgfmathprintnumber[fixed zerofill,precision=4]{\pgfplotspointmeta}},
    nodes near coords align={vertical},
    x tick label style={
        rotate=45,
        anchor=east,
        % precision=4
    },
    bar width=14pt
    ]
\addplot+[error bars/.cd,
          y dir=plus, y explicit]
coordinates {
    (No Reg, .2499) +- (0, .0003)
    (0.01\, CE, .2498) % No error bars for 'B'
    (0.05\, CE, .2531)
    (0.1\, CE, .2548)
    (0.5\, CE, .2558)
    (w/out backprop cut-off 0.01\, CE, .2496) 
    (w/out backprop cut-off 0.05\, CE, .2500) 
    (0.01\, L1, .2478) +- (0, .0024)
    (0.05\, L1, .2486)
    (0.1\, L1, .2488) 
    (0.5\, L1, .2491) 
};
\end{axis}
\end{tikzpicture}
\end{figure}

\begin{figure}[H]
\begin{center}
\begin{tikzpicture}
\begin{axis}[
     title={Performance Effect of Regularization on Peptides-Struct \textbf{(w/out Positional Encoding)}},
     width=10cm,
     ybar,
     enlargelimits=0.15,
     enlarge x limits={true},
     ylabel={Mean Absolute Loss},
     xlabel={Regularization Hyperparameters},
     symbolic x coords={
        Baseline,
        0.01 L1,
        0.05 L1,
        0.1 L1,
        0.2 L1,
        0.5 L1,
        1.0 L1,
        2.0 L1
     },
     xtick=data,
     nodes near coords={\pgfmathprintnumber[fixed zerofill,precision=4]{\pgfplotspointmeta}},
     nodes near coords align={vertical},
     x tick label style={rotate=45, anchor=east},
     ymin=0.255, ymax=0.260,
     ytick={0.255, 0.256, 0.257, 0.258, 0.259, 0.260},
     yticklabel style={
         /pgf/number format/.cd,
         fixed,               % Use fixed point arithmetic (no scientific notation)
         fixed zerofill,      % Fill numbers with zeros to ensure uniform decimal places
         precision=4,         % Number of digits after the decimal point
     },
     bar width=17pt,
    error bars/y dir=both,
    error bars/y explicit
 ]

 \addplot+[
        error bars/.cd,
        y dir=both,
        y explicit] 
coordinates {
     (Baseline,0.259) -=(0, .001) +=(0, .0001)
     % (No Positional Encoding,0.259) +- (0, .001) 
     (0.01 L1,0.2579)
     (0.05 L1,0.258)
     (0.1 L1,0.2584)
     (0.2 L1,0.2577)
     (0.5 L1,0.2555) +=(0, .0015)
     (1.0 L1,0.2571)
     (2.0 L1,0.257)
 };
 \end{axis}
 \end{tikzpicture}
% Results: Impact of Replacing Positional Encoding
 \end{center}
 \end{figure}
 
% \begin{tikzpicture}
% \begin{axis}[
%     ybar,
%     enlargelimits=0.15,
%     legend style={at={(0.5,-0.15)},
%       anchor=north,legend columns=-1},
%     ylabel={Value},
%     symbolic x coords={No reg,Sigmoid 0.01 CE,Sigmoid 0.05 CE,Sigmoid 0.1 CE,Sigmoid w/o backprop 0.01 CE,Sigmoid w/o backprop 0.05 CE,Sigmoid 0.01 L1,Sigmoid 0.05 L1,Sigmoid 0.1 L1},
%     xtick=data,
%     nodes near coords,
%     nodes near coords align={vertical},
%     x tick label style={rotate=45,anchor=east},
%     ]
% \addplot coordinates {(No reg,0.2499) (Sigmoid 0.01 CE,0.2498) (Sigmoid 0.05 CE,0.2531) (Sigmoid 0.1 CE,0.2548) (Sigmoid w/o backprop 0.01 CE,0.2496) (Sigmoid w/o backprop 0.05 CE,0.25) (Sigmoid 0.01 L1,0.2478) (Sigmoid 0.05 L1,0.2486) (Sigmoid 0.1 L1,0.2488)};
% \addplot coordinates {(No reg,0.25) (Sigmoid 0.01 CE,0.251) (Sigmoid 0.05 CE,0.254) (Sigmoid 0.1 CE,0.255) (Sigmoid w/o backprop 0.01 CE,0.248) (Sigmoid w/o backprop 0.05 CE,0.249) (Sigmoid 0.01 L1,0.247) (Sigmoid 0.05 L1,0.247) (Sigmoid 0.1 L1,0.247)};
% \legend{Category 1,Category 2}
% \end{axis}
% \end{tikzpicture} 
\hspace{1cm} The rest of the results (Peptides-Struct and PascalVOC-S) are included in the appendix section. We surprisingly observe that although the first metric values seemed to be promising in improving upon the traditional GraphGPS score, the second and third trial appeared otherwise. Despite the disappointing turn of events, we still observe that compared to no positional encoding, having the edge regularization does seem to stably help the model's performance.

\section{Application Study of GraphGPS}
\hspace{1cm} As a side-study, we also performed an application study of GraphGPS on a dataset that we believe long-range dependency exists. The dataset comes from a neutrino detector called a PhotoMultiplier Tube (PMT). Subatomic particles are incredibly difficult to observe directly, thus we utilize indirect methods to observe them. Namely, we can use a material called liquid scintillator which emits radiations when particles interact with it. The emitted radiations are different based on the event's location and the particles type (isotropic/non-isotropic radiations, etc) In our dataset, we perform a graph-level event reconstruction task where we use the features of each sensors to guess where the interaction between particle and liquid scintillator occured. We have around 70,000 such events where each event contains a fixed number of \textasciitilde 2200 sensors/nodes and each sensor contains 5 features (x, y, z positions, time, energy). Since the PMT is not inherently a graph, we perform a K-Nearest Neighbors (k=8) to train GNN. 

\hspace{1cm} We hypothesize that the communication between all sensors including far sensors are pivotal in order to perform an accurate prediction as we believe that the parameter (event location) of the radiation is difficult to know by only utilizing the local structure of the graph. 

% \begin{figure} [htbp]
\begin{figure}[H]
    \centering
    \includegraphics[width=0.75\linewidth]{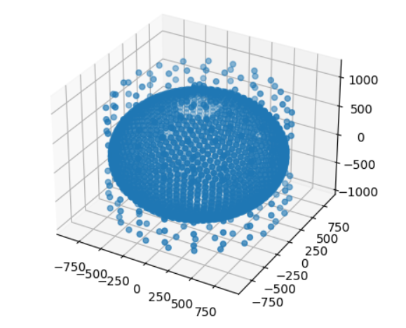}
    \caption{3D visualization of PhotoMultiplier Tube (x, y, z)}
    \label{fig:enter-label}
\end{figure}

\hspace{1cm}Before training our model, we process the data by adding addtional feature encodings: constant, clustering coefficient, Laplacian Eigenmaps(k=4), Random Walk Structural Encoding (k=16). This preprocessing have multiplied the size of the dataset by 10 times, further emphasizing the limitation of positional encodings on graph data. 

% \begin{table}[htbp]
\begin{table}[H]
\centering
\caption{Comparison between arhictectures on PMT dataset}
\begin{tabular}{|c|c|} % Two centered columns
\toprule
ARCHITECTURE & PERFORMANCE (MSE) \\
\midrule
GCN & 528 \\
GCN w/ VN & 1000\\
GraphGPS(MPNN=GCN) & 381\\
& \\
\hline
& \\
ADDED POSITIONS AS FEATURES: & \\ 
& \\
\hline
& \\
GCN & 530 \\
% GCN w/ VN & overfitted and exploded \\
GraphGPS & overfitted and exploded \\
MLP (Permutation Invariant) & somewhat-fitted and exploded \\
Transformer (Permutation Invariant) & somewhat-fitted and exploded \\
% Add more rows as needed
\bottomrule
\end{tabular}
\end{table}

\hspace{1cm}As we hypothesized, given the same number of parameters, we see that GraphGPS(MPNN(GCN) + GT) outperforms all other architectures (e.g. MPNN, MPNN+Virtual Node) we have tested. The MLP and Transformer used here were tweaked in order to obtain permutation invariance of nodes. Nevertheless, they still over-fitted. We believe that this performance evaluations provides us with some evidence that event reconstruction task from PMT dataset does indeed require long-range interaction between sensors. As for Virtual node MPNN, it is not so clear why it could not perform better than the regular MPNN. Due to time constraint, we were not able to put our regularization method to test on this dataset. In the future, we would like to see how much performance we can get out when we replace this massive feature encoding process with our edge regularization technique.

\section{Conclusion}
\hspace{1cm}In conclusion, it is not clear whether the edge regularization is a viable technique that one could use to replace positional encoding. However, it does seem evident that when no positional encoding is used, having edge regularization would likely help by a small margin. We also observe that using both positional encoding and edge regularization could interfere with each other and sometimes lead to degrade in performance. 

\hspace{1cm}However, we are optimistic since one could come up with many variations of this technique (e.g. using Shortest Path Distance regularization) and that we have not tuned any hyperparameters from the original configuration of GraphGPS. Specifically, we believe in high potentials of a clever regularization technique on the attention score matrices. 

\hspace{1cm}Despite the optimism, we still find that the memory problem of GT even by itself is quite significant. Compared to MPNN architectures, GT still struggled to fit many graphs onto GPU even without Positional Encoding. We conjecture that finding the appropriate regularization technique could possibly replace positional encoding but still will not fully resolve the GT's memory complexity problem. Lastly, we hope that the next generation of Transformer (e.g. Structured State Space Models) could perhaps alleviate this issue completely. 

\section*{References}
\begin{enumerate}
    \item Kipf, T. N., \& Welling, M. (2016). Semi-supervised classification with graph convolutional networks. arXiv preprint arXiv:1609.02907.
    \item Alon, U., \& Yahav, E. (2020). On the bottleneck of graph neural networks and its practical implications. arXiv preprint arXiv:2006.05205.
    \item Oono, K. \& Suzuki, T  (2020). Graph neural networks exponentially lose expressive power for node classification. ICLR2020, 8.
    \item Vaswani, A., Shazeer, N., Parmar, N., Uszkoreit, J., Jones, L., Gomez, A. N., ... \& Polosukhin, I. (2017). Attention is all you need. Advances in neural information processing systems, 30.
    \item Wu, Q., Zhao, W., Li, Z., Wipf, D. P., \& Yan, J. (2022). Nodeformer: A scalable graph structure learning transformer for node classification. Advances in Neural Information Processing Systems, 35, 27387-27401.
    \item Dwivedi, V. P., Rampášek, L., Galkin, M., Parviz, A., Wolf, G., Luu, A. T., \& Beaini, D. (2022). Long range graph benchmark. Advances in Neural Information Processing Systems, 35, 22326-22340.
    \item Dosovitskiy, A., Beyer, L., Kolesnikov, A., Weissenborn, D., Zhai, X., Unterthiner, T., ... \& Houlsby, N. (2020). An image is worth 16x16 words: Transformers for image recognition at scale. arXiv preprint arXiv:2010.11929.
    \item Belkin, M., \& Niyogi, P. (2003). Laplacian eigenmaps for dimensionality reduction and data representation. Neural computation, 15(6), 1373-1396.
    \item Dwivedi, V. P., Luu, A. T., Laurent, T., Bengio, Y., \& Bresson, X. (2021). Graph neural networks with learnable structural and positional representations. arXiv preprint arXiv:2110.07875.
    \item Rampášek, L., Galkin, M., Dwivedi, V. P., Luu, A. T., Wolf, G., \& Beaini, D. (2022). Recipe for a general, powerful, scalable graph transformer. Advances in Neural Information Processing Systems, 35, 14501-14515.
    \item Lim, Derek, et al. "Sign and basis invariant networks for spectral graph representation learning." arXiv preprint arXiv:2202.13013 (2022).
\end{enumerate}

\section{Appendix}
\subsection{Our Configuration of GraphGPS}
\hspace{1cm}In terms of the architecture and configuration, we use the exact same configuration of GraphGPS that was used to achieve the paper's scores. However, we had to deviate the Graph Transformer's architecture by only using a single head and taking out the "Out Projection matrix" in the Transformer. This resulted in a slight reduction in number of parameters of each architecture (40,000) and also a slight difference in the baseline from the original GraphGPS paper.

\subsection{Results of our Edge Regularization on other datasets}
\pgfplotsset{width=6cm,compat=1.9}
\begin{tikzpicture}
\begin{axis}[
    ybar,
   title={Regularization on Peptides-Func \textbf{(Both w/ \& w/out PE)}},
    width = \textwidth,
    enlarge x limits={true},
    yticklabel style={
    /pgf/number format/.cd,
    fixed,               % Use fixed point arithmetic (no scientific notation)
    fixed zerofill,      % Fill numbers with zeros to ensure uniform decimal places
    precision=4,         % Number of digits after the decimal point
    },
    ylabel={Average Precision},
    xlabel={Configuration of Regularization},
    symbolic x coords={
        Baseline, 
        0.01\, L1,
        0.05\, L1,
        0.01\, CE, 
        No-PE Baseline, 
        No-PE 0.1\, L1,
        No-PE 0.5\, L1,
        },
    xtick=data,
    nodes near coords={\pgfmathprintnumber[fixed zerofill,precision=4]{\pgfplotspointmeta}},
    nodes near coords align={vertical},
    x tick label style={
        rotate=45,
        anchor=east,
        % precision=4
    },
    bar width=18pt
    ]
\addplot+[error bars/.cd,
          y dir=both, y explicit]
coordinates {
    (Baseline, .6435)
    (0.01\, L1, .6477)
    (0.05\, L1, .6468) += (0, .0025) -= (0, .0015)
    (0.01\, CE, .6423)
    (No-PE Baseline, .5866) += (0, 0.0184)
    (No-PE 0.1\, L1, .6192) += (0, 0.0011) -= (0, .0026) 
    (No-PE 0.5\, L1, .6096) 
};
\end{axis}
\end{tikzpicture}

\hspace{1cm}On the Peptides-Func task (same dataset as Peptides-Struct), We actually observe that having our edge regularization technique stably improves upon the performance by a slight margin. As for no Positional Encoding, we indeed observe quite a significant improvement upon the baseline score without positional encoding.

\pgfplotsset{width=6cm,compat=1.9}
\begin{tikzpicture}
\begin{axis}[
    ybar,
   title={Regularization on PCQM-Contact \textbf{(Both w/ \& w/out PE)}},
    width = \textwidth,
    enlarge x limits={true},
    yticklabel style={
    /pgf/number format/.cd,
    fixed,               % Use fixed point arithmetic (no scientific notation)
    fixed zerofill,      % Fill numbers with zeros to ensure uniform decimal places
    precision=4,         % Number of digits after the decimal point
    },
    ylabel={Mean Reciprocal Rank (higher the better)},
    xlabel={Configuration of Regularization},
    symbolic x coords={
        Baseline, 
        Reg (0.01\, L1),
        Reg (0.05\, L1),
        No-PE Baseline, 
        No-PE Reg (0.1\, L1),
        No-PE Reg (0.5\, L1),
        },
    xtick=data,
    nodes near coords={\pgfmathprintnumber[fixed zerofill,precision=4]{\pgfplotspointmeta}},
    nodes near coords align={vertical},
    x tick label style={
        rotate=45,
        anchor=east,
        % precision=4
    },
    bar width=22pt
    ]
\addplot+[error bars/.cd,
          y dir=both, y explicit]
coordinates {
    (Baseline, .3389)
    (Reg (0.01\, L1), .3388)
    (Reg (0.05\, L1), .3372)
    (No-PE Baseline, .3352)
    (No-PE Reg (0.1\, L1), .3368)
    (No-PE Reg (0.5\, L1), .3376) 
};
\end{axis}
\end{tikzpicture}

\hspace{1cm}In PCQM-Contact, we observe that edge regularization with positional encoding decreases our performance, but significantly increases when positional encoding is not present.

\end{document}